\documentclass[runningheads]{llncs}
\usepackage[T1]{fontenc}
\usepackage{graphicx}
\usepackage{tabularx}
\usepackage{subcaption}
\usepackage{hyperref}

\begin{document}
\title{ReCowGnition: A Realistic Biometric Benchmark for Cow Face Recognition}
%
%
\author{Marco Huber\inst{1,2}\orcidID{0000-0003-3413-6291} \and
Marco Kiesewalter\inst{1}\and
Judith Louise Pieper\inst{3}\orcidID{0009-0008-2863-3965} \and
Bastian Kubsch \inst{3}\orcidID{0009-0001-2501-4762} \and
Naser Damer \inst{1,2}\orcidID{0000-0001-7910-7895} }

\authorrunning{Huber et al.}
%
\institute{Fraunhofer Institute for Computer Graphics Research IGD, \\ Darmstadt, Germany \and
Technische Universität Darmstadt, Darmstadt, Germany \and
Fraunhofer Institute for Computer Graphics Research IGD, \\ Rostock, Germany}

\maketitle              

\begin{abstract}
With the development of precision livestock farming and the advances in computer vision, visual animal biometrics has gained attention. Using biometric technologies that have been proven effective for humans to identify livestock can increase animal welfare as well as production efficiency. However, challenges such as complex scenarios, similar appearances, occlusions, and non-cooperative behavior, as well as the limited amount of publicly available labeled datasets, remain. In this work, we contribute a novel, publicly available cow face benchmark dataset that has been collected in a realistic automatic scenario with 6,838 images of 161 different cows at a dairy farm. In addition to the public dataset, we define two verification and four identification evaluation protocols to foster comparable research in the cow recognition research field. Further, we provide evaluation results on our dataset of six benchmark models, which include models trained on limited data, cross-species fine-tuned models, and zero-shot foundation model approaches.

\end{abstract}
\vspace{-10mm}
\section{Introduction}
\vspace{-4mm}
The growing need for high-quality meat and milk, as well as new consideration about animal welfare and sustainable farming procedures, has led to several innovations to preserve and increase food production standards \cite{DBLP:journals/corr/abs-2210-09215,MAHATO2025312}. With the development of precision livestock farming, the livestock industry has transformed from small-scale subsistence farming toward large-scale optimized and specialized grazing enterprises requiring advanced management. A major task in livestock management is the monitoring of farm animals to track physiological, phenotypical, and health characteristics over time, which requires the identification of the monitored individuals \cite{DBLP:journals/corr/abs-2210-09215}. Additionally, core components of successful cattle management, such as cattle selection, genetic improvement breeding, disease management, and medical treatment, also require the identification of individual cattle.

Traditional cattle identification technologies, such as ear tags \cite{LESLIE201086,HARMON20235043} or branding \cite{ani8080137} affect animal welfare crucially and are vulnerable to theft, loss, and fraud, and cannot ensure an effective recognition \cite{DBLP:journals/corr/abs-2210-09215}. Other solutions, such as Radio Frequency Identification (RFID) or similar sensor-based wearable systems, have proven advantages over these methods \cite{DBLP:conf/cscloud/YangLK20} but suffer from relatively high cost and limited monitoring distance, making them difficult to deploy in outdoor farming environments, where environmental conditions, animal mobility, and infrastructure constraints further reduce their reliability. Additionally, they can also be lost, damaged, or stolen.

Another technology that gained attention to identify individual cattle is automatic visual cattle recognition \cite{DBLP:journals/corr/abs-2210-09215}. Identifying cattle using biometric technologies that have proven to be effective in identifying humans in varying conditions allows leveraging and transferring existing knowledge to another application area.
The usage of principles of biometrics, such as treating identification as a feature extraction and comparison problem rather than a classification problem, also fits the use case of animal recognition. Typical biometric systems do not need to be re-trained when encountering newly bred or acquired cattle individuals, as the features can still be extracted and compared, highlighting the benefits of biometric workflow over traditional image classification. Several unique and stable biometric modalities have been proposed to be used for visual cattle identification, such as face \cite{DBLP:conf/acmturc/YaoHLLKG19,DBLP:journals/cea/XuWGCLCW22}, body, or muzzle \cite{DBLP:journals/tbbis/KumarS25a}. Of the above-mentioned methods, cattle face recognition (FR) offers the advantage that it can be performed from a high distance in a non-intrusive way with a normal camera by utilizing distinct facial features and skin texture patterns.
Several works proposed cow FR in the past; however, none of them evaluated on a shared evaluation set or following a predefined protocol, which makes tracking advances impossible. The lack of publicly available benchmark datasets has been identified as an obstacle \cite{DBLP:journals/corr/abs-2210-09215} in the area of cow FR.
Additionally, current approach heavily rely on small in-house dataset as large-scale cow face dataset have not been released yet. From a biometric perspective, it is also important to note that a majority of the works did not follow established biometric verification and identification protocols but treated the problem as a classification problem, which limits its usability in a practical scenario. 

In this paper, we contribute a new annotated cow dataset, ReCowGnition, containing a total of 6,838 images of 161 different cows that have been captured in a realistic scenario at a dairy farm on multiple sessions. This dataset is made public along with predefined evaluation protocols. We also developed a cow face detection and alignment pipeline anchored on the cow muzzle, which we also made publicly available. Additionally, we provide six different evaluation protocols, enabling future benchmarking of novel solutions in the area of cow FR. Furthermore, we provide benchmark evaluations following the proposed protocols using six different solutions of three different categories, trained from scratch with limited data, fine-tuned cross-species, and zero-shot use of foundation models, all highlighting the challenges of realistic cow FR. The dataset, the protocols, and the detection pipeline are available at: \url{https://github.com/marcohuber/recowgnition/}.

\vspace{-3mm}
\section{Related Work}
\vspace{-2mm}

\subsection{Cattle Face Evaluation Datasets}
\vspace{-1mm}
To the best of our knowledge, there is no established evaluation benchmark that allows a fair comparison between different approaches to cattle FR. So far, existing approaches have been evaluated on small private datasets from the same farm as the training data. All works used different evaluation protocols, preventing a meaningful comparison between different approaches and training datasets. The majority of related works treated cattle FR as a classification problem, limiting its practical usability in other farms. A comparison of the used evaluation datasets and our contributed dataset and evaluation scenario is provided in Table \ref{tab:rw}.

In \cite{DBLP:conf/acmturc/YaoHLLKG19}, a dataset of dairy cow face images has been collected over a period of two months using mobile devices. In total, they obtained 18,231 images covering three scenarios, frontal, side face, and occluded faces, and used 3,646 images for evaluation. The dataset is not publicly available, but the authors shared a subset of the dataset upon request. CattleFaceNet \cite{DBLP:journals/cea/XuWGCLCW22} was evaluated in a classification setup on a private dataset that was recorded using a frontal camera. The testing set only contained 232 images of 9 individual cows, and the images were collected in a single session. Another private, small-scale dataset has been utilized by Wang et al. \cite{Wang_2020}. For testing purposes, 122 pictures of 11 different cows of the Simmental breed had been extracted from videos taken by a smartphone. In \cite{DBLP:conf/ccbr/YangXCLKG19}, a private dataset of 85,200 cow face images of 1000 different Holstein dairy cows has been extracted from videos recorded with mobile devices. Out of these, 3,190 images have been used for testing purposes in a classification setup. Chen et al. \cite{DBLP:conf/icpr/ChenWZY20} collected two private datasets, with one being taken in a controlled environment and one in the wild. In total, 6089 images of 274 cows have been taken, and one half of the images of each cow are used for training while the other half is used for testing. They evaluate using an identification setup; however, since their testing setup did not ensure identity separation between training and testing data, the results might be positively biased. In \cite{DBLP:journals/cea/WengMLZZG22}, a dataset has been collected in two sessions of 50 Simmental cattle and 80 Holstein cows, with 140 images per cow and with different angles. Out of this dataset, 2,730 images were designated as testing data, and the results are evaluated as classification. Li et al. \cite{DBLP:journals/cea/LiLL22} collected 295 videos of 103 different Simmental cattle and extracted, in the end, 3,071 cow face images for testing purposes. While the dataset was originally publicly available, it is not anymore. Another private dataset was collected by Yang et al. \cite{YANG2024512} of 2,376 face images of 110 individual cows using a mobile phone camera, and they used different subsets for evaluating the influence of different factors on the identification accuracy. Xu et al. \cite{DBLP:journals/asc/XuDWZS24} also created a private dataset consisting of images of Holstein cows and utilized 1,108 images for evaluation purposes. For their research, Bergman et al. \cite{BERGMAN2024101079} collected a private dataset consisting of 7,032 images of 77 Holstein cows, and 1,230 of these images were used for evaluation. In addition to these datasets used in a scientific context, \cite{csce873cv-pd9an_dataset} provides a public dataset with incomplete identity labels. We use this dataset with manually added identity labels to train our cow face detector and cow FR system.  

As outlined above, all of the related work evaluates on a private testing split of their collected data with a limited size. With a different number of individual cows and different amounts of total images and images per cow, the achieved results are hardly comparable, making a assignment of the current state-of-the-art impossible. Additionally, sharing the same recording situation between training and test data might not represent a practical scenario, as the achieved results might not transfer to other farms or recording scenarios. To encourage evaluating in a reproducible, comparable way, we propose our novel ReCowGnition benchmark.

\begin{table}[t]
\centering
\scriptsize
\setlength{\tabcolsep}{4pt}
\caption{Overview of used evaluation scenarios and data in cow FR. Our dataset nearly doubles the total amount of currently used evaluation images and is made public.  C = Classification, V = Verification, I = Identification}
\begin{tabularx}{\linewidth}{l c ll c c}
Paper & Scenario & Eval. Data & Breed  & Public? \\
\hline \hline
Xu et al. \cite{DBLP:journals/cea/XuWGCLCW22} & C & 232 images, 9 cows & ? & No \\
Yao et al. \cite{DBLP:conf/acmturc/YaoHLLKG19} & C & 3,646 images, ? cows & ? & No \\
Wang et al. \cite{Wang_2020} & V & 122 images, 11 cows & Simmental & No \\
Yang et al. \cite{DBLP:conf/ccbr/YangXCLKG19} & C & 3,189 images, 1,000 cows & Holstein & No\\
Chen et al. \cite{DBLP:conf/icpr/ChenWZY20} & I & 3,045 images, 274 cows & ? & No \\
Weng et al. \cite{DBLP:journals/cea/WengMLZZG22} & C & 2,730 images, 130 cows & Simmental, Holstein & No \\
Li et al. \cite{DBLP:journals/cea/LiLL22} & C & 3,071 images, 103 cows & Simmental & Was\\
Yang et al. \cite{YANG2024512} & I & 1,650 images, ? cows & ? & No\\
Xu et al. \cite{DBLP:journals/asc/XuDWZS24} & C & 1,108 images, 118 cows & Holstein & No\\
Bakhshayeshi et al. \cite{DBLP:journals/iotj/BakhshayeshiETEBA24} & C, I & 375 images, 50 cows & Holstein, Jersey & Yes\\
Bergman et al. \cite{BERGMAN2024101079}  & C & 1,230 images, 77 cows & Holstein & No \\ \hline
Ours & I, V & 6,838 images, 161 cows & Holstein  & Yes \\
\end{tabularx}
\label{tab:rw}
\vspace{-6mm}
\end{table}

\vspace{-3mm}
\subsection{Biometric Face Recognition for Cattle}
\vspace{-2mm}

In \cite{DBLP:conf/acmturc/YaoHLLKG19}, a cow detection model based on Fast R-CNN is utilized, and a cattle FR model based on a classification network obtained with Progressive Neural Architecture Search \cite{DBLP:conf/eccv/LiuZNSHLFYHM18} is proposed. The approach is evaluated in a classification setup, also investigating the impact of frontal face view, side face view, and occlusions.
CattleFaceNet \cite{DBLP:journals/cea/XuWGCLCW22} is based on a fine-tuned RetinaFace to detect the cow faces combined with a MobileNet-based classification model trained with different FR losses, such as ArcFace \cite{DBLP:conf/cvpr/DengGXZ19} loss. Since it is developed and evaluated in a classification setup, the provided solution is not scalable as it has to be re-trained when new cows appear \cite{DBLP:journals/cea/XuWGCLCW22}. Wang et al. \cite{Wang_2020} proposed to use a VGG-16 network in combination with parameter transfer based on parameters obtained by training on human faces of the VGGFace dataset, leveraging the similarity of the human FR and the cattle FR problem. They evaluate in a verification scenario. In \cite{DBLP:conf/ccbr/YangXCLKG19}, Yang et al. tackle the problem of low-resolution cattle FR and propose to apply a super-resolution network to increase the resolution of cow faces. For the classification of the cows, they train a ResNet-50 model using a private dataset. To perform cattle FR on Angus cattle, Chen et al. \cite{DBLP:conf/icpr/ChenWZY20} first used a pre-trained Mask-R-CNN to detect the cow faces and then trained three different deep learning models on their private dataset and evaluated an identification scenario. However, they did not ensure identity separation between the training and testing data. In \cite{DBLP:journals/cea/WengMLZZG22}, Weng et al. approach the problem of posture change in cattle FR with a two-branch classification CNN that takes two cow face images with different angles to predict the correct cattle. Li et al. \cite{DBLP:journals/cea/LiLL22} proposed a lightweight solution for cattle FR based on a custom CNN architecture and performed classification on a private cattle dataset. In \cite{YANG2024512}, Yang et al. proposed a solution based on triplet loss and using RetinaFace as the detection algorithm. They investigated different identification scenarios, such as occluded cow faces and different color patterns on the face. An approach based on MobileFaceNet was proposed by Xu et al \cite{DBLP:journals/asc/XuDWZS24}. They utilized MobileFaceNet with additional modules for embedding enhancement and optimization to account for images with low identity information. Bergman et al. \cite{BERGMAN2024101079} evaluated several deep learning based approaches, including Vision Transformers, to identify dairy cows simultaneously using a mixture of identification and classification approaches. 

As outlined above, there is a large variety in proposed solutions, and so far, no approach has been compared with existing works, limiting meaningful comparison between approaches and a clear understanding of the state-of-the-art. Additionally, there is no established evaluation protocol, and several solutions treat identification as a classification problem, which is usually not the case in biometric recognition. 
To approach these issues, we propose, in addition to our realistic, public dataset, fixed evaluation protocols and benchmark results, which enable future research to benchmark new approaches consistently against established baselines.

\begin{figure*}
\vspace{-4mm}
    \centering
    \includegraphics[width=0.7\linewidth]{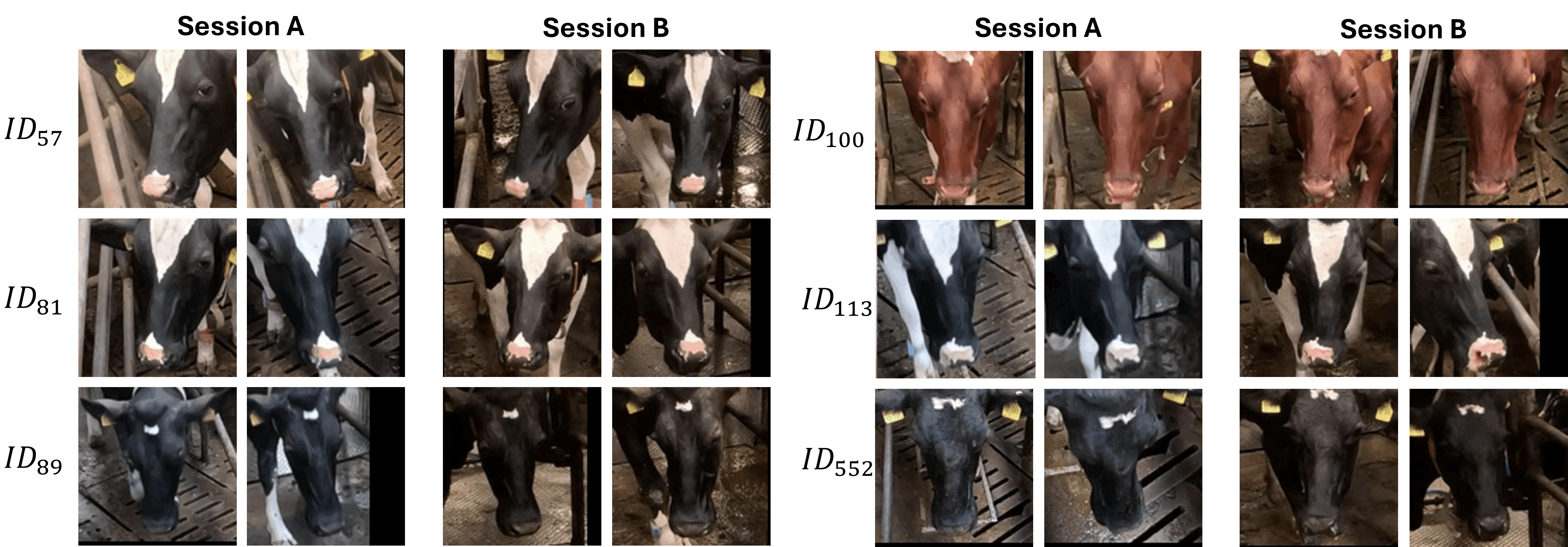}
    \caption{Example images from the ReCowGnition dataset. The images show six different identities with two images from two different sessions. The image shows the variation in pose, but also the variation in cow facial features and coat pattern.}
    \label{fig:examples}
    \vspace{-6mm}
\end{figure*} 

\vspace{-7mm}
\section{ReCowGnition: A Realistic Cow Face Dataset}
\vspace{-3mm}
In this section, we describe our cattle FR benchmark in more detail. We start with describing the video acquisition processes and the cow face detection system, CowDetect, which we developed to obtain the final images. Then, we describe the statistics of the dataset in more detail and elaborate on the challenges of cattle FR. Some example images of our dataset are shown in Figure \ref{fig:examples}.

\vspace{-3mm}
\subsection{Video Acquisition}
\begin{figure*}
\vspace{-3mm}
    \centering
    \includegraphics[width=0.7\linewidth]{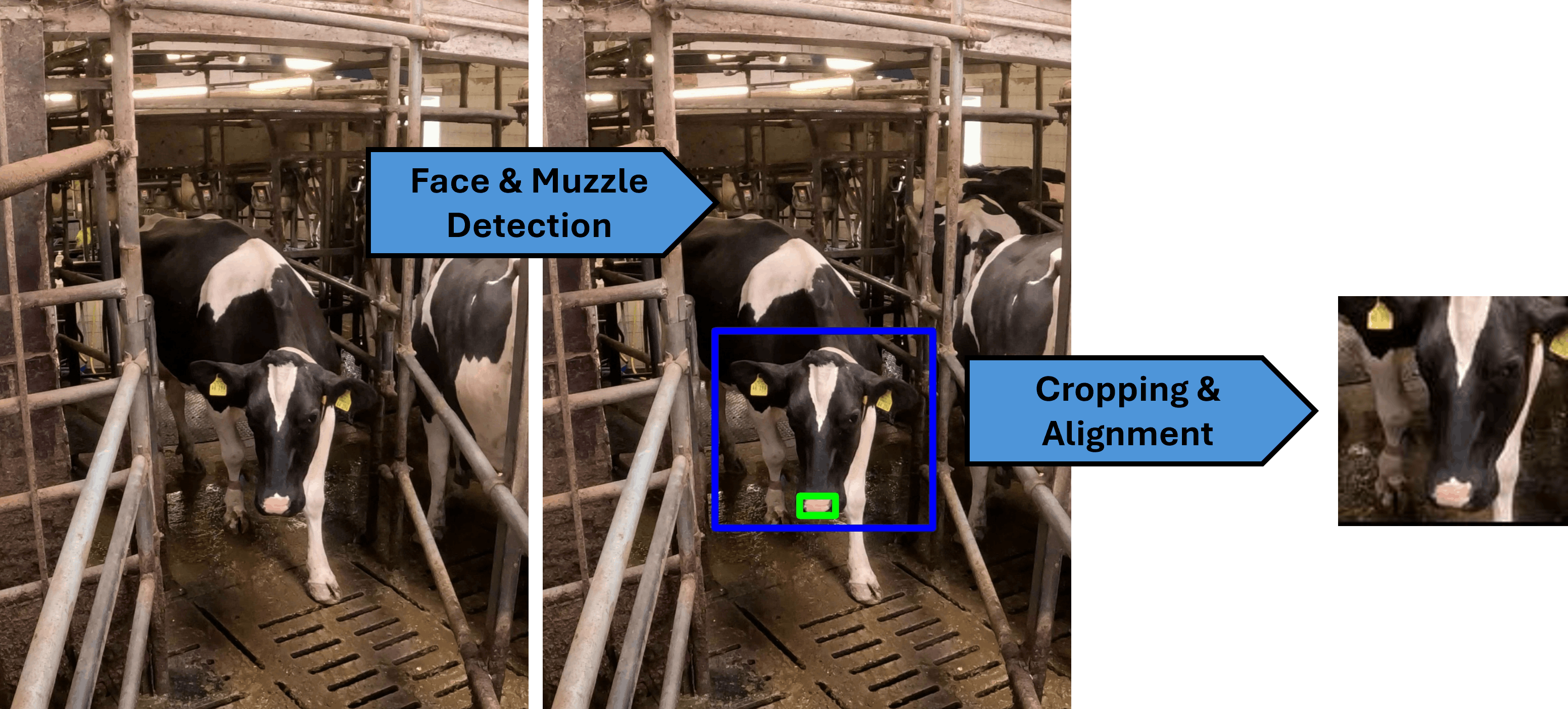}
    \caption{Cow Face Detection Pipeline: First, frames are extracted from the recorded videos. Then, CowDetect detects the cow face and muzzle. Based on the detected face and muzzle, the image is aligned and cropped.}
    \label{fig:detection}
    \vspace{-10mm}
\end{figure*}

\vspace{-2mm}
The original videos were recorded at a dairy cattle farm in Germany directly after the milking process of the cows in five recording sessions on three different days using a GoPro Hero11 Black. The camera was mounted at the exit of a walkway that the animals had to pass through, and recorded the videos autonomously. Since cows proved to be curious animals, the camera was placed at a distance and above the usual reach of a cow. The recordings took place at different times of the day 
and the camera position varied slightly for each session to increase variety and reflect different camera locations. All the cows are Holstein Friesian cows, and some of the cows are related by blood, as the majority have been bred at the farm. The age range of the cows covers 2 to 7 years, and as the cows are used for milking, all of them are female. In total, 287 videos were recorded of 161 different cows. The videos were identity labeled by an employee of the dairy farm with an identity based on the records of the dairy cow farm and identity markers that are not visible in the videos. Due to privacy reasons, as working farmers can be seen in the background, these videos will not be made publicly available. An example frame of the recorded video can be seen in Figure \ref{fig:detection}. 

\vspace{-2mm}
\subsection{Image Preprocessing \& Cow Detection System}
\vspace{-2mm}
To obtain a reasonable amount of cow images for testing purposes and to ensure variation between the images, every 5th frame has been extracted from the video streams. After the frame extraction, some of the videos have been rotated based on the recording angle to normalize the head position for the cow face detection system.  

\begin{figure}
    \centering
    \vspace{-4mm}
    \includegraphics[width=0.68\linewidth]{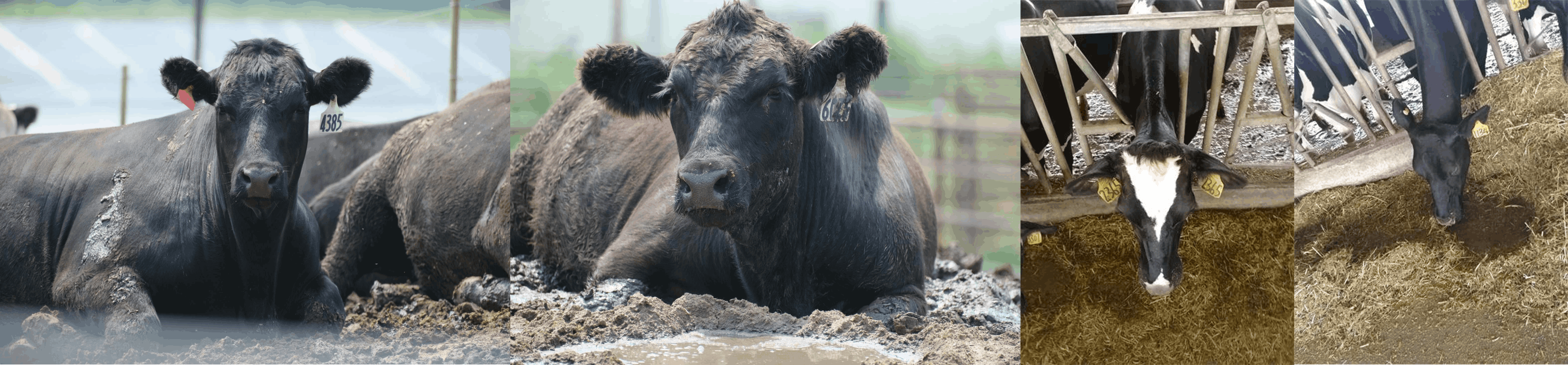}
    \caption{Example images of the training dataset for the face detection and the cow recognition models (before pre-processing): CSCE873CV \cite{csce873cv-pd9an_dataset} (left) and \cite{DBLP:conf/acmturc/YaoHLLKG19} (right)}
    \label{fig:data}
    \vspace{-10mm}
\end{figure}

To extract and align the cow face images, a new cow face detection system, CowDetect, has been developed. CowDetect is based on YOLO-v11n \cite{Jocher_Ultralytics_YOLO_2023} and has been trained to detect cow faces and cow muzzles. The detected muzzle is used to align the cow face images by ensuring that the detected muzzles always appear at the same position. This is done as face alignment has been proven as a simple technique to increase the performance of FR systems for humans \cite{DBLP:journals/corr/ZhangZL016}. Usually, for human faces, the eye, nose, and mouth positions are used for alignment, but since there is a lack of cow landmark detectors, we utilized existing muzzle labels and also manually labeled additional data. Starting from a YOLO-v11n model pre-trained on the COCO dataset \cite{DBLP:conf/eccv/LinMBHPRDZ14}, the model was fine-tuned on a subset (6,375 images) of the cow dataset used in \cite{DBLP:conf/acmturc/YaoHLLKG19}, directly provided by the authors and the publicly available CSCE873CV dataset \cite{csce873cv-pd9an_dataset}. As mentioned in the related work section, this dataset is publicly available but only partially labeled with identity labels. To use both of these for training our detection model, we manually labeled missing cow faces and muzzles. Example images of the training dataset are shown in Figure \ref{fig:data}. After augmentation, which included cropping, rotation, blur, noise, and cutout, a total of 17,299 images were used for training and 1,911 non-augmented images for testing. The model was trained for 200 epochs with a weight decay of 0.001. The achieved performance in terms of mean average precision at 50\% intersection over union (mAP@50), mean average precision averaged over IoU threshold 0.50 to 0.95 (mAP@50-95), precision, recall, and F1-score are reported in Table \ref{tab:cowdetect}. The detection test dataset is a subset of left-out images of both, \cite{csce873cv-pd9an_dataset} and \cite{DBLP:conf/acmturc/YaoHLLKG19}.

\begin{table}[t]
\centering
\scriptsize
\caption{CowDetect detection performance. The fine-tuned YOLO-v11n \cite{Jocher_Ultralytics_YOLO_2023} model is able to detect cow faces and muzzles.}
\label{tab:cowdetect}
\begin{tabular}{l|ccccc}
\hline
 & mAP@50 & mAP@50--95 & Precision & Recall & F1\\
\hline
CowDetect & 0.938 & 0.724 & 0.944 & 0.988 & 0.965 \\
\hline
\end{tabular}
\vspace{-6mm}
\end{table}

After the detection and alignment of the cow face, the image has been scaled and cropped to the size of 112x112, which is an established standard image size for human faces in the recognition purpose \cite{DBLP:conf/cvpr/DengGXZ19,DBLP:conf/cvpr/BoutrosDKK22}. Afterwards, the obtained images have been manually scanned, and mis-detected images have been removed. It is important to note that after the pre-processing, no other information that might be used for identification (such as an ear tag number) is consistently visible or readable, restricting the usable features to the cow's face features. This is important as the dataset aims to benchmark solutions that can be used in scenarios without eartags. 

\vspace{-2mm}
\subsection{ReCowGnition Benchmark Statistics \& Challenges}
\begin{figure}[htbp]
  \centering
  \begin{subfigure}{0.25\textwidth}
    \centering
    \includegraphics[width=\linewidth]{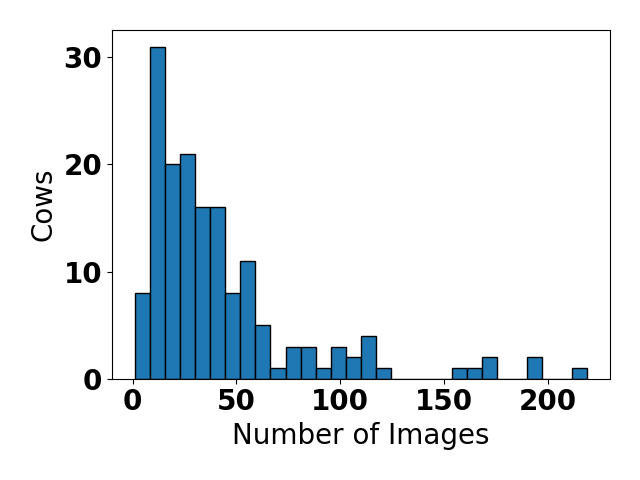}
    \caption{Images per cow}
    \label{fig:dis}
  \end{subfigure}
  \vspace{-2mm}
  \begin{subfigure}{0.25\textwidth}
    \centering
    \includegraphics[width=\linewidth]{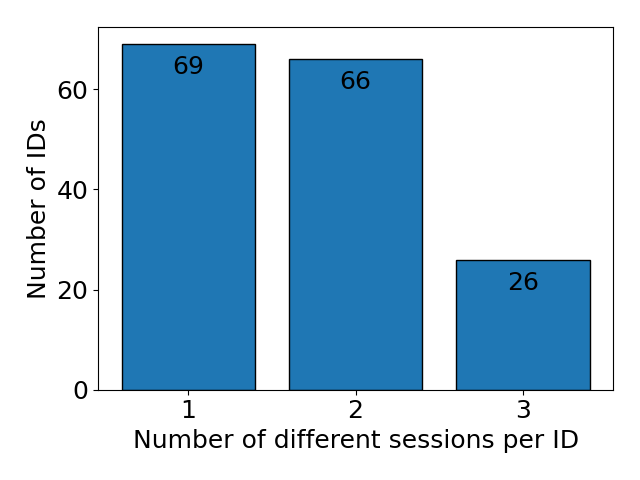}
    \caption{Sessions per cow}
    \label{fig:sess}
  \end{subfigure}
  \caption{Statistics of the ReCowGnition benchmark}
  \label{fig:both}
  \vspace{-7mm}
\end{figure}

\vspace{-2mm}
The pre-processing of the recorded videos led to 6,838 different images of 161 different cows. The distribution of the amount of images per cow is shown in Figure \ref{fig:dis}. The average amount of images per individual cow is 42 images. The amount of different sessions per cow is shown in Figure \ref{fig:sess}.

We also want to highlight some challenges and special characteristics that are present in our benchmark. The first is breeding and lineage. As selective breeding of dairy cows is a central characteristic of dairy farms to improve economically and biologically important traits, including milk production, disease resistance, fertility, and longevity, this results in many cows being related to varying degrees. While in humans it is well established that certain face features are inherited from parent generations, similar patterns can also be observed in cows \cite{https://doi.org/10.1111/age.12249}. A large number of related individuals can make recognition more difficult, as the diversity of facial features may be reduced. In our dataset, several of the cows are related to different degrees; however, this information is not tracked. The second challenge we want to highlight is the greater degree of pitch freedom enabled by the cow’s long neck. This anatomical feature allows the animal to tilt its head much further than humans typically do, resulting in facial views from steep top-down or bottom-up angles that are uncommon in human faces. Since the images were taken autonomously without human intervention on the cows' natural behavior, several steep top-down and bottom-up images are included in our realistic benchmark.

\vspace{-3mm}
\section{Evaluation Protocols}
\vspace{-3mm}
In this section, we describe the evaluation setup in more detail. In general, protocols differentiate between verification and identification. In contrast to most existing work, we propose to evaluate and approach cattle FR similarly to human FR as a biometric problem, which involves feature extraction and comparison instead of a pure classification problem. 

Approaching FR as a classification problem has the disadvantage that it requires several instances of the respective class and that these classes have to be fixed in advance before the model is trained. If a new cow is added to the farm, the model has to be updated as unknown classes appear, and not updating the model would lead to false classification, which makes this approach less practical, especially when confronted with a large amount of changing classes (individuals). Since acquiring data from all humans (or all that might approach or use a biometric system) is, in practical cases, impossible, feature extraction and comparison have been established as the gold standard in human FR. In this scenario, the model is trained to classify instances of cows but is used to extract a mathematical representation (the embedding) describing the identity features of the individual in a way that they differ between individuals. These extracted features (reference embedding) are then stored together with some identifier during the first time the individual appears (enrollment). Whenever the same individual is presented to the system, the system extracts features again (probe embedding), and the probe and reference embeddings are compared. Based on some system threshold, the model then decides if, based on the similarity of the embeddings, the individual is the same individual or not. In contrast to classification, this approach does not need any data from the individual during training, and can also be applied to other cattle populations without re-training as long as the model can still extract meaningful identity embeddings. In biometric recognition, there are usually two operation modes, which are also covered in our evaluation protocol: verification and identification. The code to evaluate following the different protocols is released along with the dataset. 

\vspace{-3mm}
\subsection{Verification Protocols}
\vspace{-2mm}
Biometric verification refers to the 1:1 comparison of a reference and a probe image of an individual that answers the question: Is this the same individual? In the area of cow FR, a possible practical scenario would be, e.g., a veterinarian who verifies that the presented cow is the one that needs to be vaccinated. For this, a probe image (e.g., taken with a smartphone) is compared to the reference image of the cow that is supposed to be vaccinated, and the system evaluates if it is the supposed cow or not.

To evaluate the verification performance, we follow the standard \cite{ISO19795-1:2021} established in human FR, which is to report the Equal Error Rate (EER) and the False Non-Match Rate (FNMR) at a certain False Match Rate (FMR). The EER is the system's performance at the decision threshold where FNMR=FMR. The FNMR@FMR metric allows to evaluate at a specific threshold that sets a different bar on the amount of false matches the systems made. To get this, the corresponding similarity threshold is determined that achieves a certain FMR, and the FNMR at this threshold is reported. This allows to value the false matches more since usually wrong matches are considered worse than false non-matches. Vaccinating the wrong cow might be worse than manually re-checking the cow. In our protocol, we report the FNMR@FMR=10\% and the FNMR@FMR=1\%. For a graphical and threshold-independent evaluation, we also provide the Receiver Operating Characteristic (ROC) Curve based on the FMR and 1-FNMR. The ROC curve illustrates the trade-off between FMR and FNMR across different decision thresholds. The combination of EER, FNMR@FMR, and ROC provides a comprehensive evaluation framework, allowing both threshold-dependent and threshold-independent performance assessment for verification.

Since we acquired data from five different sessions, we provide two different verification evaluation protocols,
$V_{ALL}$ and $V_{CS}$. For both protocols, we create all possible impostor pairs based on the whole dataset, leading to 23,105,619 impostor pairs (two images of two different cows).  In the $V_{ALL}$ protocol, all possible genuine pairs (two images of the same cow) are investigated and used for the evaluation, leading to 270,084 investigated genuine pairs. In the $V_{CS}$ protocol, we provide a harder, even more realistic evaluation scenario and remove all same-session genuine pairs. This removes easier matches originating from the same video or capture session, while same-session impostor pairs are retained. As a result, the number of genuine pairs is reduced to 90,063.

\vspace{-2mm}
\subsection{Identification Protocols}
\vspace{-2mm}
Biometric identification refers to the 1:Many comparisons of a probe image to a whole set of reference images with the goal to answer the question: Which individual is it? In the area of cow FR, a possible closed-set scenario is to identify a cow that has just been milked to connect her with the amount of milk she produced. To identify a cow, the image of the cow is taken, embedded, and then compared with all stored reference images, and the image with the highest similarity in the reference dataset is assumed to be the identity of the cow. 

To evaluate the identification performance, the Top-1 and Top-5 accuracy are reported. Top-1 accuracy measures the percentage of query images for which the highest-scoring identity prediction matches the correct cow identity. Top-5 accuracy measures the percentage of query images for which the correct identity appears among the five highest-scoring identity predictions. While Top-1 accuracy reflects the system’s ability to make a single correct identification, Top-5 accuracy provides insight into its robustness. Since not all identities have multiple images or sessions, cows with only a single image or session are not used as probes, but are still included in the gallery. In addition to the Top-1 and Top-5 accuracy, we also provide the Cumulative Match Characteristic (CMC) curve as a graphical representation of the identification performance. The CMC curve reports the probability that the correct identity appears within the top-$k$ ranked matches, as a function of the rank $k$. It therefore summarizes the identification accuracy across different rank thresholds and provides a more comprehensive view of system performance.

For the evaluation of identification performance on the ReCowGnition dataset, we propose four different identification protocols, $I_{ALL}$, $I_{CS}$, $I_{EF}$, and $I_{SF}$. In the $I_{ALL}$ evaluation setup, each image is used as a probe image (as long as the individual has more than one image, a correct identification is possible), and all other images are used as the gallery. In the $I_{CS}$ setup, same-session images are removed from the gallery based on the probe image to provide a harder, more realistic scenario. 

In contrast to this single-image approach, we also provide more complex evaluation setups with $I_{EF}$ and $I_{SF}$, which are fusion-based. In the embedding-fusion setup, $I_{EF}$, we group all embeddings from one video and calculate the mean embedding as a single representation for the identity shown in the video. This is done for the probe video and for all gallery videos, and the identification is then performed based on the fusion-based embedding. In the score-fusion approach $I_{SF}$, we also approach identification video-based and group all embeddings from one video. But instead of taking the mean, we calculate the pairwise similarities between the probe group and all gallery groups, and the mean similarity score is used for identification.

\vspace{-2mm}
\section{Benchmark Models}
\vspace{-2mm}
In addition to the benchmark evaluation dataset and the proposed evaluation protocols, we also provide results on our benchmark using six benchmark models to allow future work to compare with established approaches. These benchmark models cover six established approaches and tackle the problem of limited training data. All of the models have been evaluated on the benchmark datasets following exactly the evaluation protocol to ensure reproducibility. In the following subsections, we will outline the utilized approaches and the training setup. As the training dataset, we used the same dataset that was used to fine-tune the cow face detection system. The small-scale dataset is a combination of the CSCE873CV \cite{csce873cv-pd9an_dataset} dataset and a subset of the dataset used in \cite{DBLP:conf/acmturc/YaoHLLKG19} and consists of only 10.987 images of 571 different cows. It is also important to highlight that the training images do not cover the same individual cows (as they were recorded at different farms) and are also in a different domain (outdoor vs. indoor, feeding vs. walking) and cattle races. These conditions provide a realistic scenario for a practical and generalizable cattle FR system. 

\vspace{-4mm}
\subsection{Models trained from Scratch}
\vspace{-2mm}
The benchmark models, named ArcFace and ElasticFace-Arc, follow the intuitive approach of training a cattle FR model from scratch without any special additions neglecting the fact that data is limited. To align the architecture with the model size of the fine-tuned human-based models, we select ResNet-100 as the backbone and choose ArcFace loss \cite{DBLP:conf/cvpr/DengGXZ19} or ElasticFace-Arc loss \cite{DBLP:conf/cvpr/BoutrosDKK22} as the loss function, respectively. Both of them are well-established FR loss functions in the human domain and incorporate a margin parameter to improve the separability of the learned representations. For the margin $m$ and scaling $s$ parameter, we use the same value as reported in the original papers, which are for both: $m=0.5, s=64$. We train the models for 100 epochs with a batch size of 64 and a learning rate of $0.0125$ using Stochastic Gradient Descent with a momentum of $0.9$ and a weight decay of $0.0005$. We set the embedding size to 512 as this has been established as a standard in embedding sizes for human FR. For augmentations, we utilize random horizontal flip, Gaussian blur, and randomly adjusted sharpness. We especially did not use rotation or similar transformations as the training images are aligned based on the muzzle, similar to the testing data. We evaluate every fifth epoch and at the end, we report the performance of the best achieving model in terms of $I_{ALL}$. 

\vspace{-4mm}
\subsection{Fine-tuning Human-based Models}
\vspace{-2mm}
The benchmark models ArcFace$_{FT}$ and ElasticFace-Arc$_{FT}$ are human FR models fine-tuned for cow FR. Rather than being trained from scratch, both models were initialized from weights pre-trained on the MS1MV2 dataset \cite{DBLP:conf/cvpr/DengGXZ19}, leveraging representations learned from large-scale human face data. The same loss function (ArcFace loss \cite{DBLP:conf/cvpr/DengGXZ19} and ElasticFace-Arc loss \cite{DBLP:conf/cvpr/BoutrosDKK22}) was retained during fine-tuning, ensuring consistency between the human- and cow-based training stages. Apart from the initialization, the training setup exactly matches that of the models trained from scratch (see above, Arcface and ElasticFace-Arc), and all layers are fine-tuned. A similar fine-tuning strategy based on leveraging human face data has been proposed by \cite{Wang_2020}. This approach tackles the problem of limited training data by domain/species transfer.

\vspace{-4mm}
\subsection{Zero-Shot Foundation Models}
\vspace{-2mm}
As the last  approach, we utilize foundation models in a zero-shot setting assuming a no data scenario. For this, we employ the ViT-B/16 and ViT-L/14 models from CLIP \cite{DBLP:conf/icml/RadfordKHRGASAM21}. These models are pretrained on large-scale image–text pairs and provide generalizable image embeddings. Despite its general-purpose design, testing CLIP in a zero-shot setup allows us to assess how well foundation models capture the fine-grained differences and similarities of cow faces. This is also motivated by the fact that these models have proven to show far from random performance on human FR, even in a zero-shot approach \cite{DBLP:journals/ivc/ChettaouiDB25}. Since no model fine-tuning is performed, the system relies solely on the pretrained visual representations to compare images. The models ViT-B/16 and ViT-L/14 differ in size (base and large) and patch resolution (16 and 14)  with ViT-L/14 being larger and using smaller patches.

\vspace{-3mm}
\section{Benchmark Results}
\vspace{-3mm}
This section presents and discusses the results achieved by the benchmark models on our new benchmark dataset, ReCowGnition, in the different protocols. In Table \ref{tab:results}, the results in term of EER, FNMR@FMR, and identification accuracy are reported. 

\begin{table*}[]
\vspace{-6mm}
\centering
\resizebox{\textwidth}{!}{%
\begin{tabular}{c|rrr|rrr||rr|rr|rr|rr}
                           & \multicolumn{6}{c||}{Verification} & \multicolumn{8}{c}{Identification [in \%]} \\
                           & \multicolumn{3}{c|}{$V_{ALL}$} & \multicolumn{3}{c||}{$V_{CS}$} & \multicolumn{2}{c|}{$I_{ALL}$} & \multicolumn{2}{c|}{$I_{CS}$} & \multicolumn{2}{c|}{$I_{EF}$} & \multicolumn{2}{c}{$I_{SF}$} \\
Model                      & EER & FNMR10\% & FNMR1\% & EER & FNMR10\% & FNMR1\% & Top-1 & Top-5 & Top-1 & Top-5 & Top-1 & Top-5 & Top-1 & Top-5 \\
\hline
ArcFace                      & 0.254 & 0.518 & 0.818 & 0.305 & 0.711 & 0.957 & 82.76 & 94.09 & 9.30 & 18.15 & 8.10 & 20.95 & 5.71 & 19.05 \\
ElasticFace-Arc                      & 0.243 & 0.478 & 0.792 & 0.311 & 0.694 & 0.936 & 81.41 & 93.05 & 8.53 & 17.95 & 9.05 & 18.57 & 5.71 & 20.95 \\ \hline
ArcFace$_{FT}$                      & 0.137 & 0.194 & 0.585 & 0.163 & 0.284 & 0.742 & 95.58 & 98.87 & 30.28 & 52.58 & 34.29 & 62.38 & 30.48 & 55.71 \\
ElasticFace-Arc$_{FT}$                 & 0.129 & 0.172 & 0.533 & 0.154 & 0.260 & 0.713 & 96.39 & 99.20 & 30.30 & 53.54 & 32.86 & 61.43 & 32.86 & 60.95 \\ \hline
ViT-B/16 \cite{DBLP:conf/icml/RadfordKHRGASAM21}             & 0.364 & 0.626 & 0.842 & 0.479 & 0.873 & 0.985 & 69.23 & 86.18 & 2.29 & 8.05 & 3.33 & 8.10 & 1.43 & 6.67\\
ViT-L/14 \cite{DBLP:conf/icml/RadfordKHRGASAM21}                & 0.367 & 0.650 & 0.850 & 0.465 & 0.867 & 0.985 & 65.82 & 83.53 & 2.15 & 7.24 & 1.90 & 8.57 & 0.48 & 4.76 \\
\end{tabular}%
}
\caption{Performance of the benchmark models on the proposed ReCowGnition dataset following the defined protocols. The results show the benefit of leveraging human-based data (as seen in the performance of the fine-tuned models) and the further need of more specialized approaches to cow FR.}
\label{tab:results}
\vspace{-10mm}
\end{table*}

\begin{figure}[htbp]
  \centering
  \begin{subfigure}{0.20\textwidth}
    \centering
    \includegraphics[width=\linewidth]{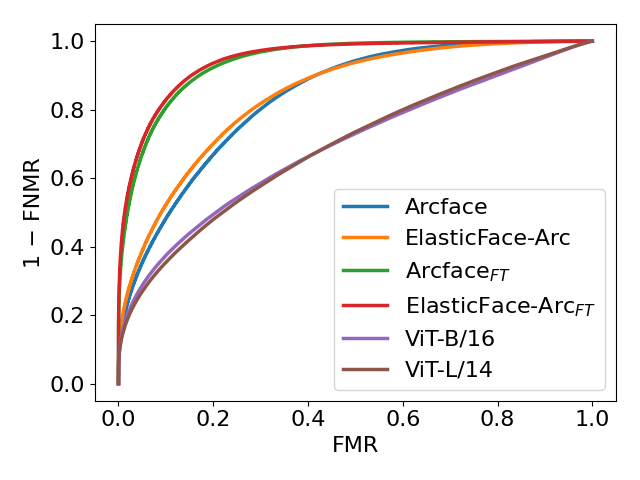}
    \caption{$V_{ALL}$-ROC}
    \label{fig:one}
  \end{subfigure}
  \begin{subfigure}{0.20\textwidth}
    \centering
    \includegraphics[width=\linewidth]{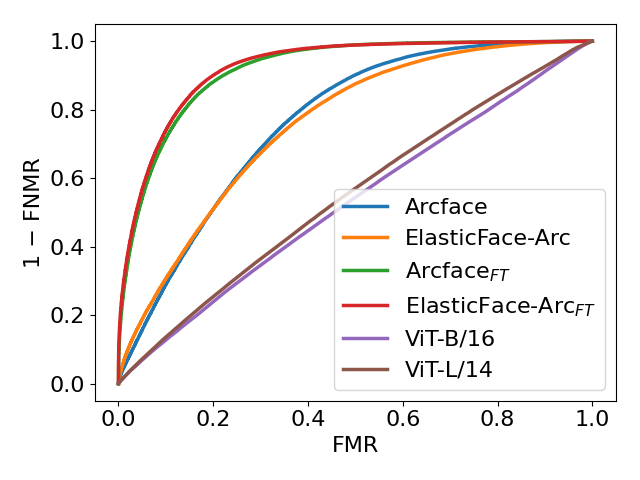}
    \caption{$V_{CS}$-ROC}
    \label{fig:two}
  \end{subfigure}
  \caption{ROC curves for the two verification protocols. The fine-tuned models outperform the other models in terms of verification performance.}
  \label{fig:roc}
  \vspace{-4.5mm}
\end{figure}

In terms of verification performance, the lowest EER was achieved by ElasticFace-Arc$_{FT}$ in both protocols, $V_{ALL}$ (0.129) and $V_{CS}$ (0.154). The superiority of the models fine-tuned in comparison to the models trained from scratch shows the benefit of leveraging pre-trained human-based FR models for cow FR. The low performance of the zero-shot foundation models, ViT-B/16 and ViT-L/14 shows that, despite their ability to extract meaningful features in various scenarios, the complex models are inadequate for verifying cows, at least in the zero-shot setup. When comparing the performance between the two protocols, all models achieved better performance in the $V_{ALL}$ evaluation than in the  $V_{CS}$ evaluation (lower EER and lower FNMR@FMR, e.g. Arcface$_{FT}$ (EER): $V_{ALL} = 0.137$ and $V_{CS}=0.164$), highlighting that removing image pairs from the same recording increases the complexity of the verification problem. The results for a graphical and threshold-independent verification performance evaluation based on ROC curves are provided in Figure \ref{fig:roc}. As illustrated in the figure, the observations from the threshold-dependent evaluation remain valid under the threshold-independent evaluation, and the fine-tuned models outperform the other models. 

\begin{figure*}[htbp]
  \centering
  \begin{subfigure}{0.20\textwidth}
    \centering
    \includegraphics[width=\linewidth]{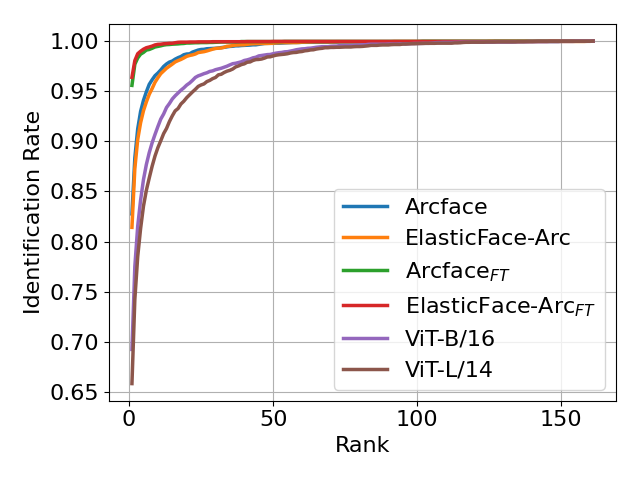}
    \caption{$I_{ALL}$-CMC}
    \label{fig:one}
  \end{subfigure}
  \begin{subfigure}{0.20\textwidth}
    \centering
    \includegraphics[width=\linewidth]{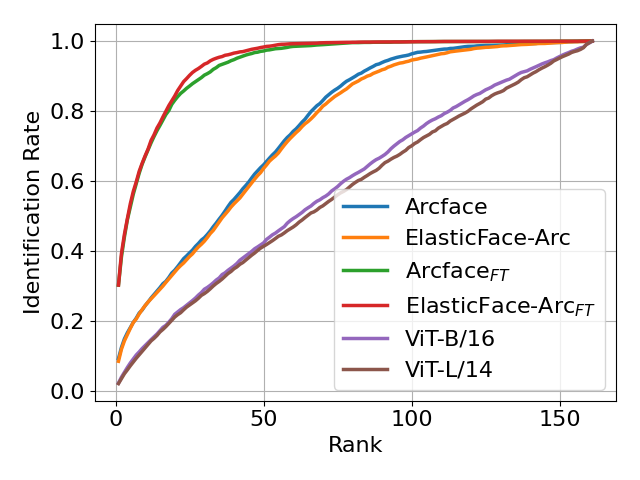}
    \caption{$I_{CS}$-CMC}
    \label{fig:two}
  \end{subfigure}
    \begin{subfigure}{0.20\textwidth}
    \centering
    \includegraphics[width=\linewidth]{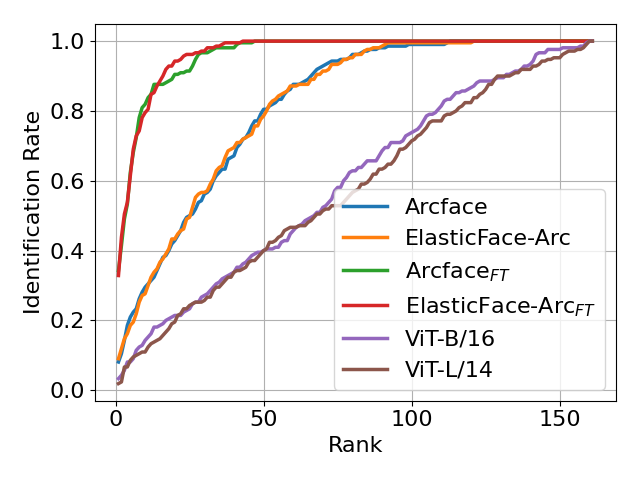}
    \caption{$I_{EF}$-CMC}
    \label{fig:one}
  \end{subfigure}
  \begin{subfigure}{0.20\textwidth}
    \centering
    \includegraphics[width=\linewidth]{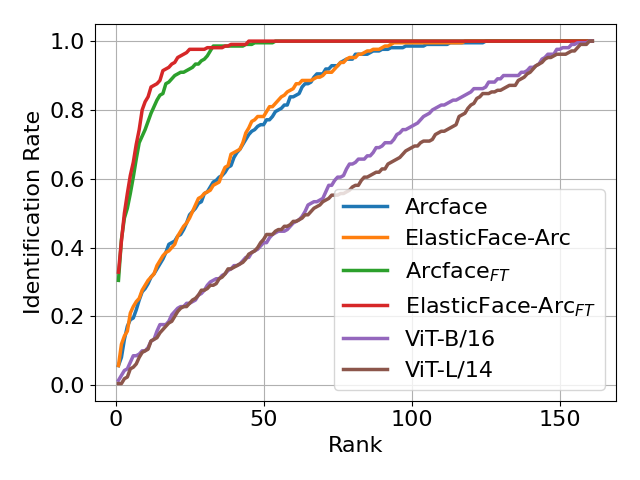}
    \caption{$I_{SF}$-CMC}
    \label{fig:two}
  \end{subfigure}
  \caption{CMC curves for the four identification protocols. Removing same-session images increases the difficulty of the problem but using fusion-strategies such as embedding or score fusion can increase the performance.}
  \label{fig:cmc}
  \vspace{-7mm}
\end{figure*}

In terms of identification accuracy, the results show similar results in terms of verification accuracy. The fine-tuned models outperform the models trained from scratch and also the zero-shot ViT models with ElasticFace-Arc$_{FT}$, achieving a Top-1 accuracy of 96.39\% following the $I_{ALL}$ protocol and a Top-1 accuracy of 30.30\% following the $I_{CS}$. This again highlights the increased complexity when images are from different recordings, motivating further research to tackle this problem and enabling reliable cow FR. While the zero-shot foundation models did not perform as competitively as the fine-tuned models, they outperformed random accuracy (0.68\% due to 161 identities) in general, and only using score-fusion with the ViT-L/14 was not beneficial. 
When comparing the results of the fusion-based evaluation scenarios, $I_{EF}$ and $I_{SF}$, which consider the whole video rather than single images, an improvement can be observed at least for the fine-tuned models (e.g. Arcface$_{FT}$, $I_{CS}$-Top-1: 30.28\% to $I_{EF}$-Top-1: 34.28\%). This shows that incorporating more information obtained from multiple images into the identification process is beneficial, especially when the model has captured more identity information. In Figure \ref{fig:cmc}, the CMC curves are provided. The curves also emphasize that removing same-session images from the gallery increases the complexity of the problem and that using fusion strategies can act as one approach to increase the performance. 

The results of the benchmark models on our new benchmark show how difficult the problem of cow FR is and that it is necessary to develop new, specialized approaches. It is crucial to benchmark new approaches on a public benchmark with a fixed protocol to enable fair and reproducible comparison. Our public dataset and the protocols we defined are intended to cover a wide range of use cases (verification, identification, and identification with fusion) and motivate future research to compare in a reproducible manner. The proposed benchmark models were evaluated to provide a baseline evaluation and to justify the selected protocols and metrics. 

\vspace{-4mm}
\section{Conclusion}
\vspace{-3mm}
Biometric cattle recognition is a crucial task in modern farm management to ensure animal welfare and efficiency; however, current approaches are limited to evaluating on small private datasets and with no established evaluation scenarios or protocols. With this work, we provide a novel evaluation benchmark, ReCowGnition, which establishes a realistic dairy farm cattle FR scenario. We provide 6,838 cow face images of 161 different cows collected in an automatic, authentic real-life scenario, a cow face detection system, as well as two verification and four identification protocols to enable comparable and reproducible research. On our new benchmark dataset, we evaluated six benchmark models belonging to three different solution families to show the complexity of realistic and practical cattle face recognition, motivating future advances and comparable research in this field.  
\vspace{-4mm}
\subsubsection*{Acknowledgement}
This research was funded through the Fraunhofer Initiative “Biobased Value Creation and Smart Farming (BWSF)” by the German Federal Ministry for Research, Technology and Space and the Ministry of Science, Culture, Federal and European Affairs of the State of Mecklenburg-Western Pomerania. This research work has also been funded by the German Federal Ministry of Education and Research and the Hessian Ministry of Higher Education, Research, Science and the Arts within their joint support of the National Research Center for Applied Cybersecurity ATHENE.
\vspace{-4mm}

\bibliographystyle{splncs04}
\bibliography{bib}

@inproceedings{DBLP:conf/acmturc/YaoHLLKG19,
  author       = {Liyao Yao and
                  Zexi Hu and
                  Caixing Liu and
                  Hanxing Liu and
                  Yingjie Kuang and
                  Yuefang Gao},
  title        = {Cow face detection and recognition based on automatic feature extraction
                  algorithm},
  booktitle    = {{ACM} {TUR-C}},
  publisher    = {{ACM}},
  year         = {2019}
}

@software{Jocher_Ultralytics_YOLO_2023,
author = {Jocher, Glenn and Qiu, Jing and Chaurasia, Ayush},
license = {AGPL-3.0},
month = jan,
title = {{Ultralytics YOLO}},
url = {https://github.com/ultralytics/ultralytics},
version = {8.0.0},
year = {2023}
}

@misc{
    csce873cv-pd9an_dataset,
    title = { CSCE873CV Dataset },
    type = { Open Source Dataset },
    author = { CattleDetection },
    howpublished = { \url{ https://universe.roboflow.com/cattledetection-dn9uy/csce873cv-pd9an } },
    journal = { Roboflow Universe },
    publisher = { Roboflow },
    year = { 2025 },
    month = { oct },
    note = { visited on 2026-01-08 },
    }

@inproceedings{DBLP:conf/eccv/LinMBHPRDZ14,
  author       = {Tsung{-}Yi Lin and
                  Michael Maire and
                  Serge J. Belongie and
                  James Hays and
                  Pietro Perona and
                  Deva Ramanan and
                  Piotr Doll{\'{a}}r and
                  C. Lawrence Zitnick},
  title        = {Microsoft {COCO:} Common Objects in Context},
  booktitle    = {{ECCV} {(5)}},
  series       = {Lecture Notes in Computer Science},
  publisher    = {Springer},
  year         = {2014}
}

@article{DBLP:journals/corr/ZhangZL016,
  author       = {Kaipeng Zhang and
                  Zhanpeng Zhang and
                  Zhifeng Li and
                  Yu Qiao},
  title        = {Joint Face Detection and Alignment using Multi-task Cascaded Convolutional
                  Networks},
  journal      = {CoRR},
  volume       = {abs/1604.02878},
  year         = {2016}
}

@article{LESLIE201086,
title = {Assessment of acute pain experienced by piglets from ear tagging, ear notching and intraperitoneal injectable transponders},
journal = {Applied Animal Behaviour Science},
volume = {127},
number = {3},
year = {2010},
issn = {0168-1591},
author = {Edwina Leslie and Marta Hernández-Jover and Ronald Newman and Patricia Holyoake}
}

@Article{ani8080137,
AUTHOR = {Adcock, Sarah J. J. and Tucker, Cassandra B. and Weerasinghe, Gayani and Rajapaksha, Eranda},
TITLE = {Branding Practices on Four Dairies in Kantale, Sri Lanka},
JOURNAL = {Animals},
VOLUME = {8},
YEAR = {2018},
NUMBER = {8},
ARTICLE-NUMBER = {137},
PubMedID = {30087245},
ISSN = {2076-2615},

}

@article{HARMON20235043,
title = {Development and application of a novel approach to scoring ear tag wounds in dairy calves},
journal = {Journal of Dairy Science},
volume = {106},
number = {7},
year = {2023},
issn = {0022-0302},
author = {Megan L. Harmon and Blair C. Downey and Alycia M. Drwencke and Cassandra B. Tucker},
}

@inproceedings{DBLP:conf/cscloud/YangLK20,
  author       = {Liuqing Yang and
                  Xiao{-}Yang Liu and
                  Jeong Soo Kim},
  title        = {Cloud-based Livestock Monitoring System Using {RFID} and Blockchain
                  Technology},
  booktitle    = {CSCloud/EdgeCom},
  publisher    = {{IEEE}},
  year         = {2020}
}

@article{DBLP:journals/cea/XuWGCLCW22,
  author       = {Beibei Xu and
                  Wensheng Wang and
                  Leifeng Guo and
                  Guipeng Chen and
                  Yongfeng Li and
                  Zhen Cao and
                  Saisai Wu},
  title        = {CattleFaceNet: {A} cattle face identification approach based on RetinaFace
                  and ArcFace loss},
  journal      = {Comput. Electron. Agric.},
  volume       = {193},
  year         = {2022}
}

@inproceedings{DBLP:conf/eccv/LiuZNSHLFYHM18,
  author       = {Chenxi Liu and
                  Barret Zoph and
                  Maxim Neumann and
                  Jonathon Shlens and
                  Wei Hua and
                  Li{-}Jia Li and
                  Li Fei{-}Fei and
                  Alan L. Yuille and
                  Jonathan Huang and
                  Kevin Murphy},
  title        = {Progressive Neural Architecture Search},
  booktitle    = {{ECCV} {(1)}},
  series       = {Lecture Notes in Computer Science},
  publisher    = {Springer},
  year         = {2018}
}

@article{Wang_2020,
year = {2020},
month = {jan},
publisher = {IOP Publishing},
volume = {1453},
number = {1},
author = {Wang, Hongyu and Qin, Junping and Hou, Qinqin and Gong, Shaofei},
title = {Cattle Face Recognition Method Based on Parameter Transfer and Deep Learning},
journal = {Journal of Physics: Conference Series},
}

@article{DBLP:journals/corr/abs-2210-09215,
  author       = {Md Ekramul Hossain and
                  Muhammad Ashad Kabir and
                  Lihong Zheng and
                  Dave L. Swain and
                  Shawn McGrath and
                  Jonathan Medway},
  title        = {A Systematic Review of Machine Learning Techniques for Cattle Identification:
                  Datasets, Methods and Future Directions},
  journal      = {CoRR},
  volume       = {abs/2210.09215},
  year         = {2022}
}

@inproceedings{DBLP:conf/icpr/ChenWZY20,
  author       = {Shunnan Chen and
                  Sen Wang and
                  Xinxin Zuo and
                  Ruigang Yang},
  title        = {Angus Cattle Recognition Using Deep Learning},
  booktitle    = {{ICPR}},
  publisher    = {{IEEE}},
  year         = {2020}
}

@inproceedings{DBLP:conf/cvpr/DengGXZ19,
  author       = {Jiankang Deng and
                  Jia Guo and
                  Niannan Xue and
                  Stefanos Zafeiriou},
  title        = {ArcFace: Additive Angular Margin Loss for Deep Face Recognition},
  booktitle    = {{CVPR}},
  publisher    = {Computer Vision Foundation / {IEEE}},
  year         = {2019}
}

@inproceedings{DBLP:conf/ccbr/YangXCLKG19,
  author       = {Zehao Yang and
                  Hao Xiong and
                  Xiaolang Chen and
                  Hanxing Liu and
                  Yingjie Kuang and
                  Yuefang Gao},
  title        = {Dairy Cow Tiny Face Recognition Based on Convolutional Neural Networks},
  booktitle    = {{CCBR}},
  series       = {Lecture Notes in Computer Science},
  publisher    = {Springer},
  year         = {2019}
}

@article{DBLP:journals/cea/WengMLZZG22,
  author       = {Zhi Weng and
                  Fansheng Meng and
                  Shaoqing Liu and
                  Yong Zhang and
                  Zhiqiang Zheng and
                  Caili Gong},
  title        = {Cattle face recognition based on a Two-Branch convolutional neural
                  network},
  journal      = {Comput. Electron. Agric.},
  volume       = {196},
  year         = {2022}
}

@article{DBLP:journals/cea/LiLL22,
  author       = {Zheng Li and
                  Xuemei Lei and
                  Shuang Liu},
  title        = {A lightweight deep learning model for cattle face recognition},
  journal      = {Comput. Electron. Agric.},
  volume       = {195},
  year         = {2022}
}

@article{YANG2024512,
title = {Fusion of RetinaFace and improved FaceNet for individual cow identification in natural scenes},
journal = {Information Processing in Agriculture},
volume = {11},
number = {4},
year = {2024},
issn = {2214-3173},
author = {Lingling Yang and Xingshi Xu and Jizheng Zhao and Huaibo Song},
}

@article{DBLP:journals/asc/XuDWZS24,
  author       = {Xingshi Xu and
                  Hongxing Deng and
                  Yunfei Wang and
                  Shujin Zhang and
                  Huaibo Song},
  title        = {Boosting cattle face recognition under uncontrolled scenes by embedding
                  enhancement and optimization},
  journal      = {Appl. Soft Comput.},
  volume       = {164},
  year         = {2024}
}

@article{BERGMAN2024101079,
title = {Biometric identification of dairy cows via real-time facial recognition},
journal = {animal},
volume = {18},
number = {3},
year = {2024},
issn = {1751-7311},
author = {N. Bergman and Y. Yitzhaky and I. Halachmi},
}

@article{MAHATO2025312,
title = {Integrating Artificial Intelligence in dairy farm management biometric facial recognition for cows},
journal = {Information Processing in Agriculture},
volume = {12},
number = {3},
year = {2025},
issn = {2214-3173},
author = {Shubhangi Mahato and Suresh Neethirajan},
}

@inproceedings{DBLP:conf/cvpr/BoutrosDKK22,
  author       = {Fadi Boutros and
                  Naser Damer and
                  Florian Kirchbuchner and
                  Arjan Kuijper},
  title        = {ElasticFace: Elastic Margin Loss for Deep Face Recognition},
  booktitle    = {{CVPR} Workshops},
  publisher    = {{IEEE}},
  year         = {2022}
}

@inproceedings{DBLP:conf/icml/RadfordKHRGASAM21,
  author       = {Alec Radford and
                  Jong Wook Kim and
                  Chris Hallacy and
                  Aditya Ramesh and
                  Gabriel Goh and
                  Sandhini Agarwal and
                  Girish Sastry and
                  Amanda Askell and
                  Pamela Mishkin and
                  Jack Clark and
                  Gretchen Krueger and
                  Ilya Sutskever},
  title        = {Learning Transferable Visual Models From Natural Language Supervision},
  booktitle    = {{ICML}},
  series       = {Proceedings of Machine Learning Research},
  publisher    = {{PMLR}},
  year         = {2021}
}

@article{DBLP:journals/tbbis/KumarS25a,
  author       = {Niraj Kumar and
                  Sanjay Kumar Singh},
  title        = {CattleDiT: {A} Distillation-Driven Transformer for Cattle Identification},
  journal      = {{IEEE} Trans. Biom. Behav. Identity Sci.},
  volume       = {7},
  number       = {4},
  year         = {2025}
}

@misc{ISO19795-1:2021,
  title        = {{ISO/IEC} 19795-1:2021(en): Information technology --- Biometric performance testing and reporting --- Part 1: Principles and framework},
  author       = {{ISO/IEC}},
  year         = {2021},
  howpublished = {International Standard}
}

@article{https://doi.org/10.1111/age.12249,
author = {Mészáros, Gábor and Petautschnig, Elisabeth and Schwarzenbacher, Hermann and Sölkner, Johann},
title = {Genomic regions influencing coat color saturation and facial markings in Fleckvieh cattle},
journal = {Animal Genetics},
volume = {46},
number = {1},
year = {2015}
}

@article{DBLP:journals/ivc/ChettaouiDB25,
  author       = {Tahar Chettaoui and
                  Naser Damer and
                  Fadi Boutros},
  title        = {FRoundation: Are foundation models ready for face recognition?},
  journal      = {Image Vis. Comput.},
  volume       = {156},
  year         = {2025}
}

@article{DBLP:journals/iotj/BakhshayeshiETEBA24,
  author       = {Ivan Bakhshayeshi and
                  Eila Erfani and
                  Firouzeh Rosa Taghikhah and
                  Stephen Elbourn and
                  Amin Beheshti and
                  Mohsen Asadnia},
  title        = {An Intelligence Cattle Reidentification System Over Transport by Siamese
                  Neural Networks and {YOLO}},
  journal      = {{IEEE} Internet Things J.},
  volume       = {11},
  number       = {2},
  year         = {2024}
}

\end{document}